\documentclass[runningheads]{llncs}

\usepackage[T1]{fontenc}
\usepackage{tikz}
\usepackage[hidelinks]{hyperref}
\usepackage{pgfgantt}
\usepackage{pdflscape}
\usepackage{capt-of} 
\usepackage{pgfplots}
\pgfplotsset{width=10cm,compat=1.9}
\usepackage{mathtools}
\usepackage{amsmath,amssymb}
\usepackage{tikz}
\usepackage{forest}
\usepackage{url}
\usetikzlibrary{trees,positioning,shapes,shadows,arrows.meta}
\usepackage{pdflscape}

\usepackage{graphicx}
\begin{document}
\title{Developing robust methods to handle missing data in real-world applications effectively\thanks{This work was presented at the ECML PKDD 2024 PhD Forum. \url{https://ecmlpkdd.org/2024/program-accepted-phd-forum/}}}
%
%

\author{Youran Zhou\orcidID{0009-0001-6831-4634}\and
Mohamed Reda Bouadjenek\orcidID{0000-0003-1807-430X} \and
Sunil Aryal\orcidID{0000-0002-6639-6824} }
\authorrunning{Y. Zhou et al.}
%
\institute{School of Information Technology, Deakin University, Geelong, Victoria, Australia \\ \mailsa}

\urldef{\mailsa}\path|{echo.zhou, reda.bouadjenek, sunil.aryal}@deakin.edu.au|

\maketitle              
\begin{abstract}
Missing data is a pervasive challenge spanning diverse data types, including tabular, sensor data, time-series, images and so on. Its origins are multifaceted, resulting in various missing mechanisms. Prior research in this field has predominantly revolved around the assumption of the Missing Completely At Random (MCAR) mechanism. However, Missing At Random (MAR) and Missing Not At Random (MNAR) mechanisms, though equally prevalent, have often remained underexplored despite their significant influence.

This PhD project presents a comprehensive research agenda designed to investigate the implications of diverse missing data mechanisms. The principal aim is to devise robust methodologies capable of effectively handling missing data while accommodating the unique characteristics of MCAR, MAR, and MNAR mechanisms. By addressing these gaps, this research contributes to an enriched understanding of the challenges posed by missing data across various industries and data modalities. It seeks to provide practical solutions that enable the effective management of missing data, empowering researchers and practitioners to leverage incomplete datasets confidently.

\keywords{missing data   \and missing mechanism \and machine learning}
\end{abstract}
\section{Introduction}
\subsection{Background}

Missing data can be characterized as the absence of values or information in particular features or attributes within a dataset. Essentially, it denotes situations where data points are either inaccessible or have not been logged for specific variables or observations. The occurrence of missing data can be traced to various factors encountered during the phases of data collection, storage, or processing. For instance, individuals participating in surveys or questionnaires may opt not to respond to particular inquiries, resulting in missing data for those specific questions. Similarly, when dealing with sensor data collected in scientific experiments, missing data can manifest if sensors malfunction or fail to accurately record data.

It's essential to recognize that missing data is a big challenge that transcends data types and domains. It can manifest in any form of data, whether it's tabular, image, text, sensor data, or other data modalities. For instance, missing values might emerge in image datasets if certain image attributes or annotations are absent or incomplete. In textual data, missing information could relate to unrecorded text segments or unavailable metadata. Thus, missing data has the ability to manifest across the data landscape, regardless of the data type or application.

The presence of missing data can be attributed to a range of underlying mechanisms, each characterized by distinct assumptions. Among these recognized mechanisms are three prominent categories: Missing Completely At Random (MCAR), Missing At Random (MAR), and Missing Not At Random (MNAR) See an example from Enders~\cite{Missing_Mechanisms}. This table \ref{tab:missing_mech} contains two variables: IQ and Job Performance Ratings. The complete data sorted by IQ is in the left two columns. Missing values for Job Performance Ratings under different mechanisms are shown in the right three columns. The symbol '?' represents a missing value in each cell. For MCAR (Missing Completely At Random) data, random rating values are missed (i.e., there is no specific mechanism governing missingness). In the MAR (Missing At Random) data, all cases with missing job performance ratings belong to participants with a lower \textit{IQ} (i.e., IQ value determines the missingness of Ratings). For MNAR (Missing Not At Random) data, all Rating values lesser than 9 are missing (i.e., some specific rating values are missing, they are not dependent on IQ values but some condition on rating itself). In these scenarios, if we simply use a single imputation method, such as the mean imputation method, the results might not be representative or could be biased. While every mechanism presents its unique challenges, special attention must be directed towards MAR and MNAR. These two mechanisms stand out as the most intricate and least transparent scenarios, lacking explicit assumptions or well-defined boundaries. Notably, they hold particular significance due to their association with sensitive data privacy concerns and the potential revelation of concealed information. The presence of all three Mechanisms introduces heightened complexity and poses formidable challenges in the accurate imputation of missing values, thereby exerting profound implications on data analysis and decision-making processes. Consequently, the exploration of relevant techniques for effectively addressing these diverse missing mechanisms becomes increasingly imperative.

\begin{table}[ht]
\centering
\begin{tabular}{|c|c||c|c|c|}
\hline
\multicolumn{2}{|c||}{\textbf{Complete Dataset}} & \multicolumn{1}{c|}{\textbf{MCAR}} & \multicolumn{1}{c|}{\textbf{MAR}}& \multicolumn{1}{c|}{\textbf{MNAR}}\\
\hline
\textbf{IQ} & \textbf{Ratings} & \textbf{Ratings} & \textbf{Ratings} & \textbf{Ratings} \\
\hline\hline
78 & 9 & ? & ? & 9 \\\hline
84 & 13 & 13 & ? & 13 \\\hline
84 & 10 & ? & ? & 10 \\\hline
85 & 8 & 8 & ? & ? \\\hline
87 & 7 & 7 & ? & ? \\\hline
91 & 7 & 7 & ? & ? \\\hline
92 & 9 & 9 & 9 & 9 \\\hline
94 & 9 & 9 & 9 & 9 \\\hline
94 & 11 & 11 & 11 & 11 \\\hline
96 & 7 & ? & 7 & ? \\\hline
99 & 7 & 7 & 7 & ? \\\hline
105 & 10 & 10 & 10 & 10 \\\hline
105 & 11 & ? & 11 & 11 \\\hline
106 & 15 & 15 & 15 & 15 \\\hline
108 & 10 & 10 & 10 & 10 \\\hline
112 & 8 & ? & 8 & ? \\\hline
113 & 12 & 12 & 12 & 12 \\\hline
115 & 14 & 14 & 14 & 14 \\\hline
118 & 16 & 16 & 16 & 16 \\\hline
134 & 12 & ? & 12 & 12 \\\hline

\end{tabular}
    \caption{Types of Missing Mechanisms}
\label{tab:missing_mech}
\end{table}

\subsection{Motivation}

In the field of data science and statistics, addressing the challenge of missing data is paramount. Real-world datasets are frequently riddled with gaps, which can lead to errors, compromised reliability of results, and hindered decision-making. Therefore, tackling these missing data challenges is pivotal for upholding the accuracy and credibility of data-driven insights.

While considerable attention has been devoted to devising solutions for handling missing values across various data types, such as tabular, time series, sensor, audio, images, textual, and video data, along with complex multimodal datasets, it is noteworthy that not all factors influencing missingness have received equal consideration. These factors encompass the missing rate, missing pattern, and the specific rules governing the missingness, referred to as missing mechanisms.

The missing rate signifies the proportion of missing values within a dataset, whereas the missing pattern pertains to discernible trends or arrangements of missing values within instances or features. Occasionally, missing patterns exhibit similarities to the underlying missing mechanisms, elucidating how and why data went missing in the first place.

However, the importance of missing mechanisms is often overlooked by researchers when handling with missing data problems. To shed light on this, we present Figure \ref{fig:MAR_MNAR_Article}, which showcases the number of articles pertaining to missing data in the Scopus database based on keyword searches. Notably, we distinguish between \textit{Special Types} and \textit{All Types} of missing data articles. The former focuses on Missing at Random (MAR) and Missing Not at Random (MNAR) data, while the latter encompasses a broader range of keywords such as \textit{missing data}, \textit{incomplete data}, and \textit{imputation}. The evident discrepancy in the number of articles dedicated to understanding missing mechanisms highlights the underexplored territory in this field.
\pgfplotsset{/pgf/number format/1000 sep=}

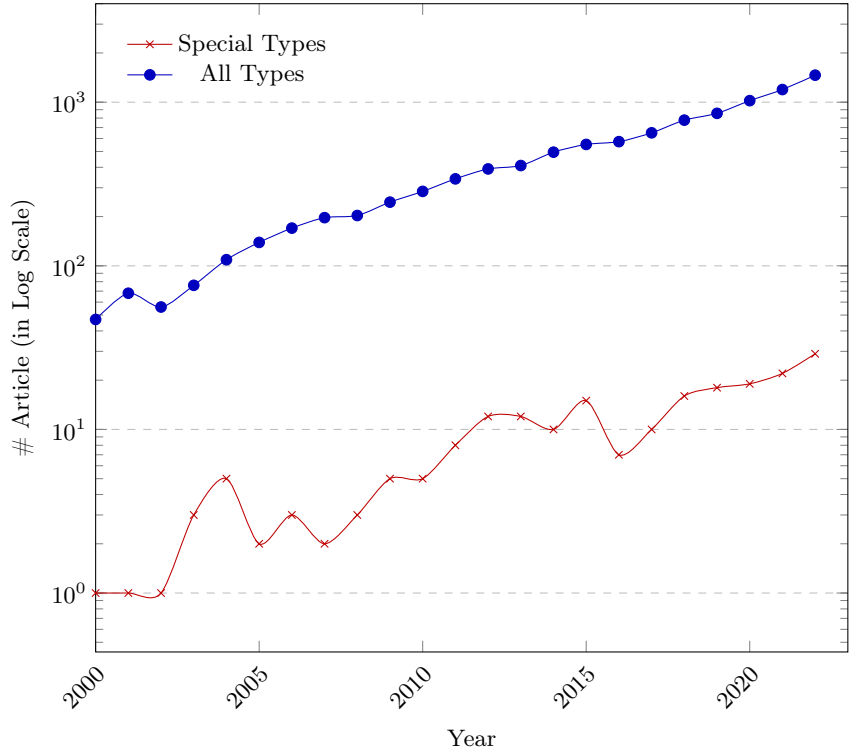
\begin{figure}[!h]
    \centering

\begin{tikzpicture}[yscale=1, xscale = 1]
    \pgfplotsset{
        scale only axis,
        y axis style/.style={
            yticklabel style=#1,
            ylabel style=#1,
            y axis line style=#1,
            ytick style=#1
       }
    }

\begin{axis}[
  ymin=0, ymax=4000,
  ymode=log,
  xmin=2000, xmax = 2023,
  xlabel=Year,
  ylabel= \# Article (in Log Scale),
  legend style={draw=none},
  xtick = {2000,2005,2010,2015,2020},
    xticklabel style={rotate=45, anchor=north east},
    legend pos=north west,
    ymajorgrids=true,
    grid style=dashed
]

\addplot[smooth,mark=x,red!75!black]
  coordinates{
(2022,	29)
(2021,	22)
(2020,	19)
(2019,	18)
(2018,	16)
(2017,	10)
(2016,	7)
(2015,	15)
(2014,  10)
(2013,	12)
(2012,	12)
(2011,	8)
(2010,	5)
(2009,	5)
(2008,	3)
(2007,	2)
(2006,	3)
(2005,	2)
(2004,	5)
(2003,	3)
(2002,1)
(2001,1)
(2000,1)
}; \label{plot_one}
\addlegendentry{Special Types}
\addplot[smooth,mark=*,blue!75!black]
  coordinates{
(2022,	1463)
(2021,	1195)
(2020,	1022)
(2019,	855)
(2018,	777)
(2017,	650)
(2016,	574)
(2015,	552)
(2014,  495)
(2013,	410)
(2012,	391)
(2011,	340)
(2010,	285)
(2009,	245)
(2008,	203)
(2007,	197)
(2006,	170)
(2005,	139)
(2004,	109)
(2003,	76)
(2002,56)
(2001,68)
(2000,47)
}; \label{plot_one}
\addlegendentry{All Types}
\end{axis}

\end{tikzpicture}

\caption{Number of Article occurs in Scopus database via Key words search}
    \label{fig:MAR_MNAR_Article}

\end{figure}

Furthermore, we also found that the majority of existing methods for handling missing data with MAR and MNAR missing mechanisms predominantly target numerical data types. This oversight presents a pressing issue, as similar challenges manifest in categorical and heterogeneous data types. Consequently, our motivation for undertaking this PhD project is born out of the necessity to bridge these critical gaps in research. By developing novel methodologies capable of addressing missing mechanisms across diverse data types, we aim to provide comprehensive solutions that empower researchers to harness the full potential of their data in an increasingly complex data landscape.

\section{Research Design}~\label{sec:ResearchDesign}
In this section, we will address the current research gap and outline the research methodology employed in this PhD project.

\subsection{Research Gap}
From our previous work  literature review \cite{zhou2024review}, the following research gap emerges:
\begin{itemize}
    \item \textbf{Scalability and Comprehensive Handling Missing Data: }\\Traditional statistical and machine learning-based methods exhibit limitations in terms of scalability and their ability to effectively handle various types of missing data. These limitations become particularly pronounced when dealing with complex missing mechanisms like missing at random (MAR) and missing not at random (MNAR).
    \item \textbf{Limited Applicability of Deep Learning:} \\Deep learning and optimization-based methods, while showing promise in addressing MAR and MNAR scenarios, are often confined to handling numerical data types exclusively. This restricts their applicability to diverse data modalities, including categorical, mixed and heterogeneous data.
    \item \textbf{Limited Applicability of Representation Learning:}\\ Representation learning methods for dealing with missing data, although versatile in accommodating multiple data types, do not adequately address the intricacies associated with missing data mechanisms. The existing literature largely lacks comprehensive solutions for recognizing and addressing different missing mechanisms, leaving this crucial aspect under-explored.
    \item \textbf{Missing Mechanisms for Various Data Types:} \\The methodology for generating missing data demonstrates a focus on certain types of tabular data. However, there is a noticeable gap in addressing missing mechanisms for other data types, such as categorical data, and multimodal data.
\end{itemize}

\subsection{Research Objects and Methodology}~\label{sec:Methodology}

This PhD project endeavors to explore robust strategies for effectively addressing missing data in real-world applications, accommodating diverse data types and varying missing mechanisms. while this PhD research will be primarily focused on tabular data, the most common data in practical applications, we aim to look into handling missing data/modalities in multimodal data.
 By 'missing modalities,' we refer to the absence of entire data types or sources in a multi-modal dataset. This objective is an aspirational objective and will be pursued towards the end of my candidature, contingent upon the availability of time.
 The overall project aims will be achieved though the following objectives and the research approaches for achieving each of the four research objectives are elaborated upon as the following:

\begin{itemize}
    \item \textbf{Objective 1: Investigating the effectiveness of existing methods in terms of handling different types of missing mechanisms and data types.}\\
    To address the challenges posed by missing data across different mechanisms, we will embark on a comprehensive research approach. Our initial step involves conducting an in-depth literature review, drawing from a pool of reputable sources that include peer-reviewed articles and conference proceedings. We will employ precise keyword-based searches on platforms like Google Scholar and Scopus to source this literature effectively. Employing advanced content analysis techniques, we will systematically categorize emerging themes concerning data types, experimental designs, and methodological approaches.

    Subsequently, we will move forward by conducting empirical studies, employing various models and frameworks. These studies will involve a meticulous examination of different missing mechanisms using both real-world datasets from the UC Irvine Machine Learning Repository\footnote{https://archive.ics.uci.edu/} and synthetic datasets. Through the application of numerical and visual evaluation metrics, we aim to assess the utility of these methods for downstream tasks, thereby advancing our understanding of effective missing data handling techniques.\\
         
     \item \textbf{Objective 2: Developing robust models for handling diverse types of missing data by investigating and enhancing existing methods to accommodate variations in missing mechanism generation techniques}
     
Our research endeavors to develop robust models for effectively addressing missing data mechanisms. We commence our investigation by scrutinizing the methodologies employed by existing models in generating and characterizing missing data mechanisms. This analysis takes into account the variations introduced by diverse data generation techniques. Subsequently, we aim to refine these established methods to comprehensively handle all types of missing data. For instance, we plan to amalgamate the mask information, drawing inspiration from MIWAE~\cite{miwae} and Not-MIWAE~\cite{notmiwae}, into TabCSDI~\cite{tabcsdi}. While TabCSDI is a powerful diffusion-based imputation method, it currently lacks the capacity to address MNAR and MAR missing mechanisms. Our goal is to bridge this gap and expand the applicability of these methods. Our research extends its scope beyond specific data types, with a primary focus on numerical datasets.\\

    \item \textbf{Objective 3: Extending the novel methods to handle different types of missing mechanisms in categorical and heterogeneous domains.}

Next, we will investigate the manifestations of missing mechanisms in different data types. Additionally, we delve into within categorical data missing mechanisms, seeking innovative strategies to apply and manage them effectively. Furthermore, we aim to pinpoint and develop techniques that capturing the complex nature of missing mechanisms. Drawing inspiration from the concept of HI-VAE \cite{nazabal2020hetroVAE}, which is a VAE-based method that can handle heterogeneous data, and Datawig \cite{biessmann2019datawig}, an optimization-based method designed for missing data in heterogeneous types, our initial focus will be on developing a method adept at handling heterogeneous data. During this stage, we will explore how to apply missing data mechanisms within heterogeneous data settings, necessitating meticulous attention.\\

    \item \textbf{Objective 4: Extending proposed methods to handle missing modalities in multimodal data.}\\
In this phase, we will delve into the integration of missing mechanisms within multimodal data, encompassing information from various sources and modalities. Noteworthy datasets in this context include enviromental sensor dataset\footnote{https://discover.data.vic.gov.au/dataset/all-sensors-real-time-status}.

Our next step involves conducting a comprehensive investigation into the interactions and mutual influences of these mechanisms across different modalities. Particular attention will be devoted to Graph Neural Networks (GNNs) and representation learning methods. As GNNs have demonstrated their efficacy in handling multimodal datasets.

\end{itemize}

%
%
%
\bibliographystyle{splncs04}
\bibliography{ref}
%




\end{document}